\documentclass[sigconf]{acmart}

\copyrightyear{2023}
\acmYear{2023}
\setcopyright{rightsretained}
\acmConference[WSDM '23]{Proceedings of the Sixteenth ACM International Conference on Web Search and Data Mining}{February 27-March 3, 2023}{Singapore, Singapore}
\acmBooktitle{Proceedings of the Sixteenth ACM International Conference on Web Search and Data Mining (WSDM '23), February 27-March 3, 2023, Singapore, Singapore}\acmDOI{10.1145/3539597.3570435}
\acmISBN{978-1-4503-9407-9/23/02}

\AtBeginDocument{%
	\providecommand\BibTeX{{%
			\normalfont B\kern-0.5em{\scshape i\kern-0.25em b}\kern-0.8em\TeX}}}

\usepackage{flushend}
\usepackage{lineno}
\usepackage{amsmath,amsfonts}
\usepackage{algorithm}
\usepackage{algorithmic}
\usepackage{graphicx}
\usepackage{textcomp}
\usepackage{xcolor}
\usepackage{subfigure} 
\usepackage{amsthm}
\usepackage{url}
\usepackage{float}
\usepackage{multirow}
\usepackage{multicol}
\usepackage{color}
\usepackage{bm}
\usepackage{bbm}

\usepackage{pifont}

\newcommand{\X}{\mathbf{X}}
\newcommand{\bU}{\mathbf{U}}

\newcommand{\bu}{\mathbf{u}}

\newcommand{\bL}{\mathbf{L}}
\newcommand{\bD}{\mathbf{D}}

\newcommand{\bB}{\mathbf{B}}

\newcommand{\x}{\mathbf{x}}
\newcommand{\y}{\mathbf{y}}

\usepackage{array}
\newcolumntype{L}[1]{>{\raggedright\let\newline\\\arraybackslash\hspace{0pt}}m{#1}}
\newcolumntype{C}[1]{>{\centering\let\newline  \\\arraybackslash\hspace{0pt}}m{#1}}
\newcolumntype{R}[1]{>{\raggedleft\let\newline \\\arraybackslash\hspace{0pt}}m{#1}}

\usepackage{ bbold }



\begin{document}

	\title{Few-shot Node Classification with Extremely Weak Supervision}

	\author{Song Wang}
\affiliation{%
  \institution{University of Virginia} \country{}}
\email{sw3wv@virginia.edu}
	
	\author{Yushun Dong}
\affiliation{%
  \institution{University of Virginia} \country{}}
\email{yd6eb@virginia.edu}

		\author{Kaize Ding}
\affiliation{%
  \institution{Arizona State University}  \country{}}

\email{kding9@asu.edu}

				\author{Chen Chen}
\affiliation{%
  \institution{University of Virginia} \country{}}
\email{zrh6du@virginia.edu}

					\author{Jundong Li}
\affiliation{%
  \institution{University of Virginia} \country{}}
\email{jundong@virginia.edu}

	\begin{abstract}

		Few-shot node classification aims at classifying nodes with limited labeled nodes as references.
		Recent few-shot node classification methods typically learn from classes with abundant labeled nodes (i.e., meta-training classes) and then generalize to classes with limited labeled nodes (i.e., meta-test classes). Nevertheless, on real-world graphs, it is usually difficult to obtain abundant labeled nodes for many classes. In practice, each meta-training class can only consist of several labeled nodes, known as the \emph{extremely weak supervision} problem. In few-shot node classification, with extremely limited labeled nodes for meta-training, the generalization gap between meta-training and meta-test will become larger and thus lead to suboptimal performance. To tackle this issue, we study a novel problem of few-shot node classification with extremely weak supervision and propose a principled framework X-FNC under the prevalent meta-learning framework. Specifically, our goal is to accumulate meta-knowledge across different meta-training tasks with extremely weak supervision and generalize such knowledge to meta-test tasks. To address the challenges resulting from extremely scarce labeled nodes, we propose two essential modules to obtain pseudo-labeled nodes as extra references and effectively learn from extremely limited supervision information. We further conduct extensive experiments on four node classification datasets with extremely weak supervision to validate the superiority of our framework compared to the state-of-the-art baselines. 
		
	\end{abstract}

	\maketitle

\section{Introduction}
\vspace{0.05in}
Node classification focuses on learning a model that can assign labels for unlabeled nodes on a graph~\cite{kipf2017semi,huang2020graph,ding2020inductive}. Many real-world analytical tasks can be formulated as the node classification problem~\cite{liu2021relative,ding2022data}. For example, in disease diagnosis, the types of diseases are regarded as class labels while patients are represented by nodes on a patient similarity graph~\cite{liu2020heterogeneous}. Recent studies mainly leverage Graph Neural Networks (GNNs)~\cite{wu2020comprehensive,velivckovic2017graph} to learn node representations and classify unlabeled nodes based on the learned presentations. However, GNNs typically require a considerable number of labeled nodes for all classes to learn effective node representations~\cite{zhou2019meta,ding2020graph}. In practice, it is often difficult to obtain sufficient labeled nodes for each class as the labeling process requires a lot of human efforts~\cite{ding2021weakly,wang2022faith}. Hence, there is a surge of research interests aiming at performing node classification with limited labeled nodes as references, known as \emph{few-shot node classification}.

	To effectively solve the few-shot node classification problem, many recent works adopt a meta-learning strategy~\cite{liu2021relative,huang2020graph,ding2020inductive}. In specific, these works learn transferable knowledge from classes with abundant labeled nodes (i.e., \emph{meta-training classes})
	and then generalize such knowledge to other classes with limited labeled nodes (i.e., \emph{meta-test classes}). The overall process is conducted on a series of meta-tasks (i.e., meta-training tasks and
	meta-test tasks), where each meta-task contains a small number of 
	\emph{support nodes} as references and several \emph{query nodes} to be classified. 
	Despite their empirical success, existing approaches~\cite{zhou2019meta,huang2020graph,ding2020graph} simply assume that meta-training classes consist of abundant labeled nodes, i.e., all the nodes in the meta-training classes are gold-labeled. However, such an assumption is generally unrealistic in practice, since each class may only consist of an extremely limited number of labeled nodes on real-world graphs. 
	For example, in molecular property prediction, certain chemical properties (i.e., classes) only consist of extremely limited labeled molecules due to the expensive cost of the labeling process~\cite{guo2021few}.
	With extremely inadequate labeled nodes for each meta-training class (i.e., extremely weak supervision~\cite{ding2022toward}), the effectiveness of the meta-learning paradigm for learning transferable meta-knowledge will be severely impacted. Thus, solving the problem of few-shot node classification with extremely limited labeled nodes for each meta-training class requires urgent research efforts. In this regard, we investigate a novel problem of \emph{few-shot node classification with extremely weak supervision} in this paper. Specifically, the goal is to perform few-shot node classification after learning from extremely limited labeled nodes for each class. 
	
\vspace{0.035in}

	However, it remains a challenging task to achieve this goal due to two major reasons. First, with extremely weak supervision, the model performance will be deteriorated by the \emph{under-generalizing}
	problem due to extremely inadequate support nodes.
	Specifically, given extremely limited labeled nodes for each meta-training class, the support nodes in each meta-training task are only sampled from a small set of labeled nodes. Therefore, the effectiveness of meta-learning in extracting meta-knowledge from different classes will be greatly weakened. As a result, the generalizability of the model to meta-test classes will drop significantly (i.e., under-generalizing). 
	Second,	with extremely weak supervision, the meta-training efficacy will be severely impacted by the \emph{over-fitting} problem due to extremely inadequate query nodes.
	Recent few-shot node classification studies~\cite{zhou2019meta,ding2020graph,ding2020inductive,liu2021relative} generally require a large number of query nodes during meta-training for model optimization. Nevertheless, 
	with extremely weak supervision from the meta-training classes, the number of query nodes for optimization during meta-training is significantly reduced. As a result, the model will be easily over-fitted and result in suboptimal performance.
	
\vspace{0.035in}
	
	To tackle the aforementioned challenges, we propose a novel framework for few-shot node classification with extremely weak supervision from the meta-training classes, named as \textit{X-FNC}. Essentially, our framework consists of two innovative modules to handle the \emph{under-generalizing} and \emph{over-fitting} issues, respectively. First, to compensate for the insufficient support nodes during meta-training, we perform label propagation to obtain abundant pseudo-labeled nodes based on Poisson Learning~\cite{calder2020poisson}. With the pseudo-labeled nodes, we can expand the support set in each meta-task to better extract discriminative meta-knowledge for each class.
	Second, to alleviate the negative impact of over-fitting caused by inadequate query nodes, we propose to optimize the model by both classifying nodes and filtering out irrelevant information
	(e.g., decisive classification information for classes not used in a meta-task) 
	based on Information Bottleneck (IB)~\cite{tishby2015deep}. As a result, in addition to learning with supervision information, the model will also learn to ignore irrelevant information during the meta-learning process, which relieves over-fitting caused by insufficient supervision information.
	 In summary, our main contributions are as follows:
	\begin{itemize}
	    \item We 
	    investigate a novel research problem of few-shot node classification with extremely weak supervision.
	    \item We develop a novel few-shot node classification framework under the extremely weak supervision scenario with two essential modules: 
	    (1) a label propagation module based on Poisson Learning to expand the support set in each meta-task by obtaining pseudo-labeled nodes; 
	    (2) an optimization strategy based on Information Bottleneck to learn from classifying query nodes while reducing irrelevant information.
	    \item We conduct extensive experiments on 
	    four node classification datasets with extremely weak supervision. Experimental results demonstrate the superiority of our framework.
	\end{itemize}

	

    \section{Related Work}	
    

    \subsection{Few-shot Node Classification}
    Few-shot learning aims to achieve considerable classification performance using limited labeled samples as references. The general approach is to accumulate transferable knowledge from tasks with abundant labeled samples and then generalize such knowledge to novel tasks with few labeled samples. Generally, there are two main categories of approaches for few-shot learning: (1) \emph{Metric-based} approaches focus on learning a metric function to match the query set with the support set for classification~\cite{liu2019learning,sung2018learning}. For example, Prototypical Networks~\cite{snell2017prototypical} learn prototypes for classes and classify query samples based on the Euclidean distances between the query set and the prototypes. (2) \emph{Optimization-based} approaches aim to optimize model parameters based on gradients on support samples in each meta-task~\cite{mishra2018simple,ravi2016optimization,nichol2018first}. As a classic example, MAML~\cite{finn2017model} learns the parameter initialization for different meta-tasks with the proposed meta-optimization strategy. 
    On graph data, 
    many research efforts have been devoted to studying few-shot learning on graphs with limited labeled nodes~\cite{tan2022graph,wang2021reform,wang2022graph,tan2022tlp}. For example, 
    Meta-GNN~\cite{zhou2019meta} combines meta-learning~\cite{finn2017model} with GNNs to reduce the requirement of labeled nodes. 
    GPN~\cite{ding2020graph} estimates node importance and leverages Prototypical Networks~\cite{snell2017prototypical} for few-shot node classification. TENT~\cite{wang2022task} proposes to reduce the variance among tasks for generalization performance.

    \subsection{Semi-supervised Few-shot Learning}
    Several recent approaches aim to combine semi-supervised or self-supervised learning with few-shot learning to improve the performance on few-shot classification tasks with unlabeled data. Ren et al.~\cite{ren2018meta} extend Prototypical Networks with unlabeled data based on the Soft k-Means method. TPN~\cite{liu2018transductive} propagates labels of given data to unlabeled data, combined with a meta-learning strategy for optimization. 
    On the other hand, the Information Bottleneck (IB) principle is also leveraged in self-supervised representation learning. DVIB~\cite{alemi2016deep} first utilizes IB in neural networks for robust representation learning. Moreover, GIB~\cite{wu2020graph} develops information-theoretic modeling of graph structures and node features on graph representation learning. 

	\section{Preliminaries}
		\subsection{Few-shot Node Classification}
	We denote an attributed graph as $\mathcal{G}=(\mathcal{V},\mathcal{E},\X)$, where $\mathcal{V}$ and $\mathcal{E}$ denote the set of nodes and edges, respectively. $\X\in\mathbb{R}^{|\mathcal{V}|\times d}$ is the node feature matrix, where $d$ is the feature dimension. Moreover, we denote the set of node classes as $\mathcal{C}$, which can be further divided into two sets: $\mathcal{C}_{train}$ and $\mathcal{C}_{test}$. Note that $\mathcal{C}=\mathcal{C}_{train}\cup\mathcal{C}_{test}$ and $\mathcal{C}_{train}\cap\mathcal{C}_{test}=\emptyset$, where $\mathcal{C}_{train}$ and $\mathcal{C}_{test}$ denote the set of meta-training and meta-test classes, respectively. General few-shot settings assume that labeled nodes in $\mathcal{C}_{train}$ are abundant, while labeled nodes in $\mathcal{C}_{test}$ are generally scarce. However, it is usually unrealistic in practice to obtain adequate labeled nodes for all classes in $\mathcal{C}_{train}$. With extremely weak supervision, the number of labeled nodes in $\mathcal{C}_{train}$ is severely limited. Subsequently, our goal is to develop a learning model such that after meta-training on extremely limited labeled nodes, the model can accurately predict labels for the nodes in $\mathcal{C}_{test}$ with only $K$ labeled nodes for each of $N$ randomly sampled classes as the reference.
	In this way, the problem is called $N$-way $K$-shot node classification. 
	

	\subsection{$N$-way $K$-shot Meta-learning}
We follow the prevalent episodic meta-learning paradigm, which has demonstrated superior performance in few-shot learning~\cite{snell2017prototypical,finn2017model,vinyals2016matching}. Particularly, we employ $\mathcal{C}_{train}$ and $\mathcal{C}_{test}$ for meta-training and meta-test, respectively. 
During meta-training, the model learns from a series of \emph{meta-training tasks}. Each meta-training task consists of a support set $\mathcal{S}$ as the reference and a query set $\mathcal{Q}$ to be classified. Here $\mathcal{S}=\{(v_1,y_1),(v_2,y_2),\dotsc,(v_{N\times K},y_{N\times K})\}$ contains $N$ classes randomly sampled from $\mathcal{C}_{train}$ and $K$ labeled nodes for each of these $N$ classes (i.e., $N$-way $K$-shot). $v_i\in\mathcal{V}$ is a node in $\mathcal{G}$ and $y_i$ is the class of $v_i$. The query set $\mathcal{Q}=\{(v^*_1,y^*_1),(v^*_2,y^*_2),\dotsc,(v^*_{Q},y^*_{Q})\}$ consists of totally $Q$ different nodes from these $N$ classes. Note that during the classification process in each meta-task, all nodes on the graph other than nodes in this meta-task are considered unlabeled and can be leveraged to advance the classification performance. During meta-test, the model is evaluated on meta-test tasks, which share a similar structure with meta-training tasks, except that the classes are in $\mathcal{C}_{test}$. Under the meta-learning~\cite{finn2017model,zhou2019meta,huang2020graph} framework, we first fine-tune the model based on support nodes and then conduct classification on query nodes.

			\begin{figure*}[htbp]
	    \centering
	    \includegraphics[width=0.85\textwidth]{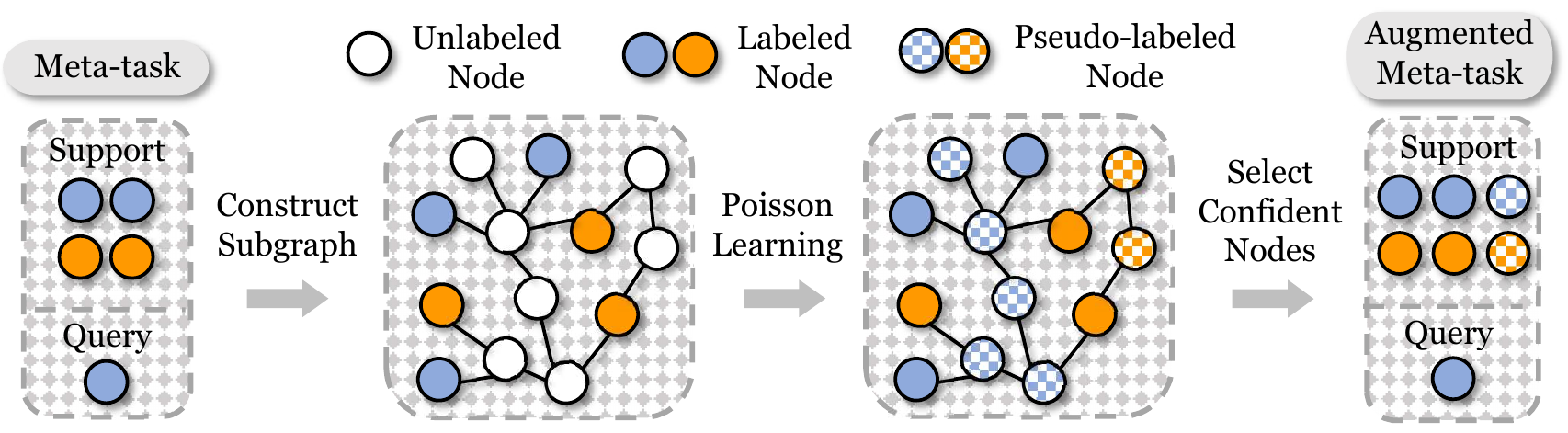}
	    \vspace{-0.08in}
\caption{The pseudo-labeling process with our Poisson Label Propagation module. For each meta-task, we construct a  subgraph based on support nodes and randomly sampled unlabeled nodes, including their neighboring nodes. Then we perform Poisson Label Propagation to obtain pseudo-labeled nodes. After that, we further select pseudo-labeled nodes with high confidence to form the augmented support set. The augmented support set will be used for fine-tuning in this meta-task.}
\label{fig:model}
\vspace{-0.1in}
	\end{figure*}

		\section{The Proposed Framework}
	
	We first present an overview of our proposed framework X-FNC. Specifically, we formulate the problem of \emph{few-shot node classification with extremely weak supervision} under the prevalent $N$-way $K$-shot meta-learning framework.
	In practice, we conduct meta-training on a series of randomly sampled meta-tasks, where a meta-task contains $K$ nodes for each of $N$ classes as the support set and several query nodes to be classified. Due to the extremely limited labeled nodes during meta-training, the model performance will be severely deteriorated by two problems: \emph{under-generalizing} (caused by inadequate support nodes) and \emph{over-fitting} (caused by inadequate query nodes). Therefore, we propose two essential modules: \emph{Poisson Label Propagation} and \emph{Information Bottleneck Fine-tuning}, which solve these two problems by obtaining pseudo-labeled nodes and maximally learning decisive information, respectively.

    \subsection{Poisson Label Propagation}
    \label{section:poisson}
    To alleviate the problem of under-generalizing caused by extremely limited support nodes during meta-training, 
    we propose to obtain pseudo-labeled nodes based on Poisson Learning~\cite{calder2020poisson}. Specifically, Poisson Learning is recently proposed to propagate labels from relatively limited labeled samples to unlabeled samples, based on the assumption that samples that are close to each other can potentially share similar classes. By recursively aggregating label information from close samples, the unlabeled samples can be pseudo-labeled based on label information propagated from labeled samples. Thus, it is helpful for obtaining pseudo-labeled nodes under the extremely weak supervision setting. However, it remains non-trivial to perform Poisson Learning on few-shot node classification with extremely weak supervision due to the following two reasons. 
    First, Poisson Learning cannot fully take advantage of structural information on graph data when leveraged to obtain pseudo-labeled nodes. Originally proposed for image classification, Poisson Learning constructs a graph solely based on the Euclidean distances between images. However, on graph data, the graph structures encode crucial information for node classification and thus cannot be ignored.
	Second, Poisson Learning cannot effectively handle a varying class set.
	Few-shot learning models are required to deal with various classes across different meta-tasks, which contradicts the fact that Poisson Learning is originally proposed to operate on a fixed class set. Nevertheless, the ability to handle various classes is crucial for few-shot node classification~\cite{ding2020graph,huang2020graph}.

	To overcome these two difficulties, as illustrated in Fig.~\ref{fig:model}, we propose to construct a subgraph in each meta-task based on graph structures and node features, and such subgraph consists of support nodes and randomly sampled unlabeled nodes. 
	In addition, we include the neighbors of these nodes and the corresponding edges in the subgraph to effectively leverage local structures of support nodes.
	Moreover, we utilize the constructed subgraph in each meta-task instead of the entire graph, so that we can perform label propagation regarding the varying classes in different meta-tasks.

    Specifically, consider a meta-task $\mathcal{T}=(\mathcal{S},\mathcal{Q})$. We first aim to sample a number of unlabeled nodes for label propagation based on Poisson Learning. Basically, the neighbors of nodes in $\mathcal{S}$ bear a higher chance of belonging to classes in $\mathcal{S}$ than other random nodes. In addition, only using these neighboring nodes can be insufficient when the average node degree is small.
    Therefore, we sample unlabeled nodes via two strategies: neighbor sampling and random sampling. For the neighbor sampling, we select 2-hop neighbors of the labeled nodes in $\mathcal{S}$, since neighboring nodes maintain explicit connections to these labeled nodes and are thus more likely to share the same classes. In particular, denoting the set of 2-hop neighbors of node $v_i$ as $\mathcal{N}_i$, the node set obtained via neighbor sampling is $\mathcal{V}_n=\bigcup_{i=1}^{NK}\mathcal{N}_i$. 
   For the random sampling, we randomly select $R$ nodes from the remaining node set $\mathcal{V}\setminus\left(\mathcal{S}\cup\mathcal{V}_n\right)$ to form a random node set $\mathcal{V}_r$, where $|\mathcal{V}_r|=R$. Then similarly, we extract the 2-hop neighbors of nodes in $\mathcal{V}_r$ as $\mathcal{V}_{rn}$. In consequence, combining nodes sampled from the two sampling strategies, we can obtain the final node set $\mathcal{V}_s=\mathcal{S}\cup\mathcal{V}_n\cup\mathcal{V}_r\cup\mathcal{V}_{rn}$ for the subgraph. 
	
	Nevertheless, the sampled nodes in $\mathcal{V}_s$ can be distributed across the entire graph and potentially unconnected, which greatly hinders the process of label propagation. Therefore, we propose to construct a subgraph with these nodes based on both the structural and feature information. More specifically, we first extract the corresponding edge set $\mathcal{E}_s$ from $\mathcal{E}$ according to $\mathcal{V}_s$. Then we denote $\mathbf{A}'\in\mathbb{R}^{|\mathcal{V}_s|\times|\mathcal{V}_s|}$ as the adjacency matrix obtained from graph structures (i.e., $\mathcal{E}_s$), where $\mathbf{A}'_{ij}=1$ if the $i$-th node in $\mathcal{V}_s$ connects to the $j$-th node in $\mathcal{V}_s$, and $\mathbf{A}'_{ij}=0$, otherwise.
	In this way, we can construct edges without losing the original structural information. Furthermore, to incorporate feature information, we propose to compute another edge weight matrix based on Euclidean distances between node features~\cite{calder2020poisson} as follows:
	\begin{equation}
	    \mathbf{A}''_{ij}=\exp\left(-\eta\|\x_i-\x_j\|\right),
	\end{equation}
	where $\eta\in\mathbb{R}$ is a hyper-parameter to control the scale of $\mathbf{A}''$ and $\|\cdot\|$ is the $\ell_2$-norm. In this way, all nodes in $\mathcal{V}_s$ are also connected according to their distances, which further advances the label propagation. Finally, we combine the two matrices to form the final adjacency matrix: $\mathbf{A}=\lambda\mathbf{A}'+(1-\lambda)\mathbf{A}''$ with a scaling hyper-parameter $\lambda\in[0,1]$. As a result, the edges can absorb information from both graph structures and node features, which effectively promotes the label propagation process based on Poisson Learning on this subgraph. Then with the learned adjacency matrix $\mathbf{A}$, we can perform Poisson Learning on this subgraph to obtain pseudo-labeled nodes.
	
    Denote $\bu_i\in\mathbb{R}^N$ as the label vector of the $i$-th node $v_i$ in $\mathcal{V}_s$ to be learned, where the index of the largest element in $\bu_i$ indicates that $v_i$ belongs to this class.
    Intuitively, Poisson Learning~\cite{calder2020poisson} assumes that the label vector of an unlabeled node is the weighted average of its neighbors' label vectors, where the weight is from the corresponding entry in $\mathbf{A}$. Moreover, the label vectors of given labeled nodes are their corresponding classes minus the average label vector of all labeled nodes. In this way, the objective of Poisson Learning can be formulated as follows:
        \begin{equation}
    \left\{
        \begin{aligned}
        \sum\nolimits_{j=1}^{|\mathcal{V}_s|} \mathbf{A}_{ij}\left(\bu_i-\bu_j\right)=0, \ &\ \ \text{if}\ \  NK+1\leq i \leq |\mathcal{V}_s|,\\
        \bu_i=\y_i-\Bar{\y}, \ &\ \ \text{if}\ \  1\leq i \leq NK,
        \end{aligned},
        \right.
        \label{eq:poisson}
    \end{equation}
       satisfying $\sum_{i=1}^{|\mathcal{V}_s|}d_i\bu_i=0$, where $d_i=\sum_{j=1}^{|\mathcal{V}_s|} \mathbf{A}_{ij}$. 
       $\y_i\in\mathbb{R}^N$, where the $j$-th element is 1 if $\x_i$ belongs to the $j$-th class, and other elements are 0. $\Bar{\mathbf{y}}=\sum_{i=1}^{NK}\y_i/NK$ is the average label vector. 
       To solve Eq.~(\ref{eq:poisson}), 
       we iteratively update the prediction matrix $\mathbf{U}\in\mathbb{R}^{\mathcal{V}_s\times N}$ based on~\cite{calder2020poisson} as follows:
        \begin{equation}
            \bU^{(t)}\leftarrow \bU^{(t-1)}+\bD^{-1}\left(\bB^\top-\bL\bU^{(t-1)}\right),
            \label{eq:poisson_iteration}
        \end{equation}
    where $t\in\{1,2,\dotsc,T_l\}$ and $T_l$ is the number of label propagation steps. $\mathbf{D}$ is a diagonal matrix and $\mathbf{D}_{ii}=\sum_{j=1}^{|\mathcal{V}_s|} \mathbf{A}_{ij}$. $\mathbf{L}=\mathbf{D}-\mathbf{A}$ is the unnormalized Laplacian matrix, and $\mathbf{B}=[\mathbf{F}-\Bar{\mathbf{y}},\mathbf{O}]$,
    where $\mathbf{O}\in\mathbb{R}^{N\times(|\mathcal{V}_s|-NK)}$ is a zero matrix. $\mathbf{F}\in\mathbb{R}^{N\times NK}$ denotes the label matrix of the $NK$ labeled nodes, whose $i$-th column is $\y_i$. The $i$-th row of the final result $\bU^{(T_l)}$ is the obtained label vector of $v_i$. The iteration is achieved by replacing the label vector (i.e., $\bu_i$) with the weighted average of label vectors from neighboring nodes of $v_i$.

	In this way, we can obtain a considerable number of pseudo-labeled nodes to compensate for the lack of support nodes during meta-training. Nevertheless, some of the pseudo-labeled nodes could be incorrect and thus deteriorate the classification performance if all pseudo-labeled nodes are used to expand the support set. Therefore, we propose to select pseudo-labeled nodes with high prediction confidence for fine-tuning in each meta-task. Specifically, we compute the confidence score for each pseudo-labeled node according to the entropy of the prediction result as follows:
	\begin{equation}
	    c_i=-\sum\limits_{j=1}^N u_{ij}\log u_{ij},
	\end{equation}
	where $u_{ij}$ is the $j$-th element of $\mathbf{u}_i$ after softmax. 
	In this way, we can select top $M$ pseudo-labeled nodes 
	with the highest confidence scores.
	Note that $\mathcal{V}_s$ contains the $NK$ support nodes, which will be ignored during the selection since they are already labeled. Then the support set can be augmented as $\widetilde{\mathcal{S}}=\mathcal{S}\cup\mathcal{S}_p$, where $\mathcal{S}_p$ is the set of selected pseudo-labeled nodes and $|\widetilde{\mathcal{S}}|=NK+M$.


			\begin{figure*}[htbp]
	    \centering
	    \includegraphics[width=0.95\textwidth]{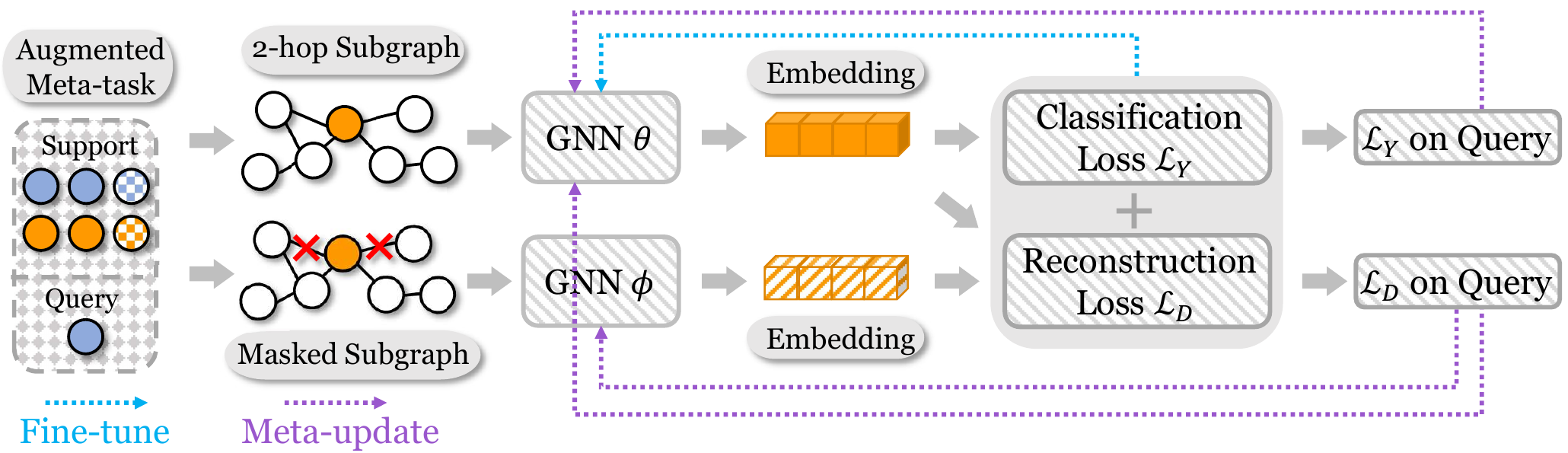}
	    \vspace{-0.1in}
\caption{The optimization process with our IB fine-tuning module and meta-learning strategy. For each node in the augmented support set, we construct a 2-hop subgraph and a masked 2-hop subgraph. Then we utilize two $\text{GNN}_\theta$ and $\text{GNN}_\phi$ to perform IB fine-tuning via $T$ steps. After that, we calculate the loss on the query set and meta-update model parameters.
}
\label{fig:optimization}
\vspace{-0.12in}
	\end{figure*}

    \subsection{Information Bottleneck Fine-tuning}
    With the augmented support set $\widetilde{\mathcal{S}}$, we can conduct fine-tuning on $\widetilde{\mathcal{S}}$ for fast adaptations to the given meta-task $\mathcal{T}$ and then meta-optimize the model on the query set $\mathcal{Q}$. However, although the support set is augmented, the query set $\mathcal{Q}$ could still be inadequate for optimization with extremely weak supervision. In other words, the model can be easily influenced by irrelevant information (e.g., 
    decisive classification information for classes not in $\mathcal{T}$
    ) 
    and thus leads to \emph{over-fitting}. 
    Therefore, we aim to fine-tune the model with extremely limited query nodes while ignoring the irrelevant information as much as possible. Particularly, the Information Bottleneck (IB)~\cite{tishby2015deep} provides an essential principle to extract classification information while maximally reducing the negative impact of irrelevant information. Moreover, the IB principle can also encourage the model to benefit from incorrect pseudo-labeled nodes by learning to neglect irrelevant information. Nevertheless, it remains non-trivial to utilize the IB principle on graph data, due to the fact that graph data does not follow the i.i.d. assumption used in previous IB-based models~\cite{wu2020graph}. Thus, we further derive two variational bounds for IB to fine-tune the model in a more tractable manner.

    
    Specifically,
    the objective of the IB principle can be formulated as follows:
    \begin{equation}
        \min\text{IB}(D,Y;Z)\triangleq[-I(Y;Z)+\beta I(D;Z)],
        \label{eq:ib}
    \end{equation}
    where $Y$ denotes the class set of nodes in $\widetilde{\mathcal{S}}$ and $|Y|=N$. $Z$ denotes the node representations to be learned. $D$ denotes the structural and feature information of nodes in $\widetilde{\mathcal{S}}$. $\beta$ is a positive scalar to balance the trade-off between the desire to preserve classification information and being invariant to irrelevant graph structures and node features~\cite{zbontar2021barlow}. In particular, the IB aims to learn representations that are maximally informative for classification (i.e., maximizing $I(Y;Z)$) while reducing irrelevant information (i.e., minimizing $I(D;Z)$). Furthermore, it becomes more useful in few-shot learning since each meta-task is only conducted on $N$ classes, and thus the irrelevant information $D$ can be more redundant. 
    
    Nevertheless, it is difficult to directly optimize the objective in Eq.~(\ref{eq:ib}), since it is intractable~\cite{wu2020graph}. Thus, we propose to derive an upper bound for each of the two terms in Eq.~(\ref{eq:ib}) for optimization. Specifically, the first term can be expressed using entropy as follows:
    \begin{equation}
        \begin{aligned}
            -I(Y;Z)
            &=-\left[H(Y)-H(Y|Z)\right],
        \end{aligned}
    \end{equation}
    where $H$ is the entropy. Since we aim to optimize the model for better $Z$, we can ignore the unrelated term $H(Y)$. Then we can obtain the explicit form of $H(Y|Z)$ based on the definition of entropy:
    \begin{equation}
        \begin{aligned}
            H(Y|Z)
            &=-\sum\limits_{i=1}^{N}\sum\limits_{j=1}^{|\widetilde{\mathcal{S}}|}p(y_i,z_j)\log\frac{p(y_i,z_j)}{p(z_j)}\\
            &=-\sum\limits_{i=1}^{N}\sum\limits_{j=1}^{|\widetilde{\mathcal{S}}|}p(z_j|y_i)p(y_i)\log p(y_i|z_j),
        \end{aligned}
    \end{equation}
    where $y_i$ and $z_i$ denote the label and the representation of the $i$-th node $v_i$ in $\widetilde{\mathcal{S}}$, respectively.
 Since each meta-task contains $K$ support nodes for each of $N$ classes,
 we can assume that the prior distribution of $Y$ is uniform, and thus $p(y_i)$ is a constant. To further estimate $p(z_j|y_i)$, we compute it via $p(z_j|y_i)=\mathbb{1}(z_j\in y_i)$, where $\mathbb{1}(z_j\in y_i)=1$ if $z_j$ belongs to $y_i$; otherwise $\mathbb{1}(z_j\in y_i)=0$. In this way, the objective of $-I(Y;Z)$ is formulated as a cross-entropy loss:
    \begin{equation}
        -I(Y;Z)\rightarrow\mathcal{L}_Y=-\sum\limits_{i=1}^{\widetilde{\mathcal{S}}}\log p(y'_i|z_i),
    \end{equation}
	where $y'_i$ denotes the specific label that the $i$-th node $v_i$ belongs to. Then to estimate $p(y'_i|z_i)$, we further utilize a $\text{GNN}_\theta$ followed by an MLP classifier $\text{MLP}_\theta$. Specifically, for the $i$-th node $v_i$ in $\widetilde{\mathcal{S}}$, we extract its 2-hop neighboring nodes to form a subgraph, represented by $(\mathbf{A}_i,\mathbf{X}_i)$. Here $\mathbf{A}_i$ and $\mathbf{X}_i$ denote the adjacency and feature matrix, respectively. Then we compute the output prediction score as
	\begin{equation}
	    \mathbf{s}_i=\text{MLP}_\theta\left(\text{GNN}_\theta(\mathbf{A}_i,\mathbf{X}_i)\right),
	\end{equation}
	where $\mathbf{s}_i\in\mathbb{R}^{N}$ is the unnormalized prediction score of $v_i$. With a softmax function, we can normalize $\mathbf{s}_i$ to finally obtain $p(y'_j|z_j)$. In this way, the model learns the crucial information for classification of classes in $Y$ via maximizing $I(Y;Z)$.
	
	For another term $I(D;Z)$, we first express it via the expectation:
	    \begin{equation}
        I(D;Z)=\mathbb{E}\left(\log\frac{p(Z|D)}{p(Z)}\right),
    \end{equation}
	where $D$ denotes the structural and feature information of nodes in $\widetilde{\mathcal{S}}$. It is noteworthy that nodes on graphs do not follow the i.i.d. assumption and thus are inherently correlated. Hence, although the fine-tuning is conducted on a specific meta-task, $D$ should incorporate the information from the entire graph due to the correlations among nodes (i.e., $D=(\mathcal{E},\mathbf{X})$). However, it is difficult to estimate $p(Z|D)$, since the only $D$ is represented by 
	the entire graph. Therefore, we introduce another distribution $q(Z)$ to approximate the true posterior $p(Z|D)$. In this way, we can further derive an upper bound of $I(D;Z)$ for optimization:
		    \begin{equation}
    \begin{aligned}
        \mathbb{E}\left(\log\left(\frac{p(Z|D)}{q(Z)}\frac{q(Z)}{p(Z)}\right)\right)
        &=\mathbb{E}\left(\log\frac{p(Z|D)}{q(Z)}\right)-\text{KL}\left(p(Z)||q(Z)\right)\\
        &\leq \mathbb{E}\left(\log\frac{p(Z|D)}{q(Z)}\right),
        \end{aligned}
    \end{equation}
    where $\text{KL}(\cdot||\cdot)$ denotes the KL-divergence of distributions. In this way, the final objective is to minimize the KL-divergence between $p(Z|D)$ and $q(Z)$. In practice, to estimate $p(Z|D)$, we utilize $\text{GNN}_\theta$ to instantiate $p(Z|D)$. However, incorporating structural and feature information from the entire graph can be inefficient and redundant for classification in a meta-task. 
    Thus, we leverage the local-dependence assumption~\cite{wu2020graph} of graph data to define $D$ as the specific structural and feature information of each node in $\widetilde{\mathcal{S}}$. In this way, $\text{GNN}_\theta$ can learn to provide a comprehensive estimation for $p(Z|D)$ based on the specific $D$ in each meta-task, since $D$ changes with $\widetilde{\mathcal{S}}$ in different meta-tasks.
    On the other hand, $q(Z)$ is a prior distribution for $Z$ and is thus difficult to estimate. Therefore, we propose to instantiate $q(Z)$ with another GNN parameterized by $\phi$ (i.e., $\text{GNN}_\phi$). Meanwhile, since $q(Z)$ is not conditioned on $D$, it is necessary to alleviate the inevitable influence of $D$. Therefore, we propose to randomly mask the corresponding graph structures and node features in the subgraph $(\mathbf{A}_i,\mathbf{X}_i)$ of each node in $\widetilde{\mathcal{S}}$. Specifically, for a subgraph represented by $(\mathbf{A}_i,\mathbf{X}_i)$, each entry in $\mathbf{A}_i$ and $\mathbf{X}_i$ has a probability of $\gamma$ to be masked (i.e., becomes zero), and the masked matrices are denoted as $(\widetilde{\mathbf{A}}_i,\widetilde{\mathbf{X}}_i)$. As a result, the model can learn to extract the decisive information for classification while maximally ignoring irrelevant information in $D$. Then for the $i$-th node $v_i$ in $\widetilde{\mathcal{S}}$, as illustrated in Fig.~\ref{fig:optimization}, we can achieve the two representations obtained by $\text{GNN}_\theta$ and $\text{GNN}_\phi$ as follows:
    \begin{equation}
        \mathbf{h}_i=\text{GNN}_\theta(\mathbf{A}_i,\mathbf{X}_i),\ \ \ 
     \widetilde{\mathbf{h}}_i=\text{GNN}_\phi(\widetilde{\mathbf{A}}_i,\widetilde{\mathbf{X}}_i),
    \end{equation}
    where $\mathbf{h}_i$ and $\widetilde{\mathbf{h}}_i$ denote the representations of $v_i$ from the two GNNs, respectively. To minimize the KL-divergence between $p(Z|D)$ and $q(Z)$,
    we utilize a predictor~\cite{grill2020bootstrap,thakoor2021bootstrapped} $p_\theta$ (a two-layer MLP) that uses $\mathbf{h}_i$ to produce a prediction $p_\theta(\mathbf{h}_i)$ for $\widetilde{\mathbf{h}}_i$. After normalizing both $p_\theta(\mathbf{h}_i)$ and $\widetilde{\mathbf{h}}_i$, the mean squared error can be defined as follows:
    \begin{equation}
        \text{MSE}(p_\theta(\mathbf{h}_i),\widetilde{\mathbf{h}}_i)= \left\|\frac{p_\theta(\mathbf{h}_i)}{\|p_\theta(\mathbf{h}_i)\|}-\frac{\widetilde{\mathbf{h}}_i}{\|\widetilde{\mathbf{h}}_i\|}\right\|^2=2-2\cdot \frac{p_\theta(\mathbf{h}_i)\cdot\widetilde{\mathbf{h}}_i}{\|p_\theta(\mathbf{h}_i)\|\|\widetilde{\mathbf{h}}_i\|},
    \end{equation}
    where $\|\cdot\|$ denotes the $\ell_2$-norm. In this way, the loss becomes:
    \begin{equation}
        I(D;Z)\rightarrow\mathcal{L}_D=-\sum\limits_{i=1}^{\widetilde{\mathcal{S}}} \frac{p_\theta(\mathbf{h}_i)\cdot\widetilde{\mathbf{h}}_i}{\|p_\theta(\mathbf{h}_i)\|\|\widetilde{\mathbf{h}}_i\|},
        \label{eq:ld}
    \end{equation}
    which is the cosine similarity between $p_\theta(\mathbf{h}_i)$ and $\widetilde{\mathbf{h}}_i$. Then the final fine-tuning loss can be defined as $\mathcal{L}=\mathcal{L}_Y+\beta\mathcal{L}_D$, where $\beta$ is the hyper-parameter in the IB principle to trade off the two mutual information terms. 
    
    
    \vspace{-0.01in}
	\subsection{Meta Learning-based Optimization}
	In this part, we elaborate on the optimization process of X-FNC. As illustrated in Fig.~\ref{fig:optimization}, our optimization process consists of two main stages: fine-tuning and meta-optimization. Given a specific meta-task $\mathcal{T}$, we first obtain the augmented support set $\widetilde{\mathcal{S}}$ via the proposed Poisson Label Propagation module introduced in Sec.~\ref{section:poisson}. Then we fine-tune our framework on $\widetilde{\mathcal{S}}$ for a fast adaptation to this meta-task. Furthermore, to ensure adaptations to each meta-task during evaluation, we utilize the prevalent strategy~\cite{finn2017model,li2019learning}, which meta-optimizes model parameters according to loss on the query set. The original strategy is proposed to optimize an entire model with one meta-learning rate. However, X-FNC consists of multiple modules with various purposes. Therefore, we propose to separately optimize modules in X-FNC based on different losses. 
	
	Specifically, let $\theta$ denote the total parameters of $\text{GNN}_\theta$, $\text{MLP}_\theta$, and the predictor $p_\theta$. 
	For the fine-tuning process, we first initialize the parameters for fine-tuning as $\theta_0\leftarrow\theta$. Then we conduct $T$ steps of fine-tuning based on the loss $\mathcal{L}$ calculated on $\widetilde{\mathcal{S}}$ as follows:
	\begin{equation}
	    \theta_t \leftarrow \theta_{t-1} -\alpha\nabla_{\theta_{t-1}}\mathcal{L}\left(\widetilde{\mathcal{S}};\theta_{t-1}\right),
	    \label{eq:fine-tuning}
	\end{equation}
	where $t\in\left\{1,2,\dotsc,T\right\}$ and $\mathcal{L}(\widetilde{\mathcal{S}};\theta_{t-1})$ denotes that the loss is calculated based on $\widetilde{\mathcal{S}}$ with the parameters $\theta_{t-1}$. $\alpha$ is the learning rate in each fine-tuning step.
	
	It is noteworthy that during fine-tuning, other parameters of our framework (i.e., $\text{GNN}_\phi$ parameterized by $\phi$) are kept unchanged, since simultaneously optimizing $\text{GNN}_\theta$ and $\text{GNN}_\phi$ can result in collapse (e.g., a constant representation)~\cite{grill2020bootstrap,thakoor2021bootstrapped}.
	After $T$ steps of fine-tuning, we will meta-optimize
	$\text{GNN}_\phi$ with the loss calculated on the query set $\mathcal{Q}$. Meanwhile, since different modules bear various purposes, we optimize them with two meta-learning rates and losses. More specifically, on the query set $\mathcal{Q}$, we meta-optimize $\theta$ and $\phi$ with the following update functions:
		\begin{equation}
	    \theta=:\theta-\beta_1\nabla_{\theta}\mathcal{L}(\mathcal{Q};\theta_T),\ \ \ 
	    \phi=:\phi-\beta_2\nabla_{\phi}\mathcal{L}_D(\mathcal{Q};\theta_T),
	    \label{eq:meta_update2}
	\end{equation}
    where $\beta_1$ and $\beta_2$ are meta-learning rates for $\theta$ and $\phi$, respectively. Note that $\text{GNN}_\phi$ is only used to calculate $\mathcal{L}_D$. Thus, $\text{GNN}_\phi$ will be meta-optimized regarding $\mathcal{L}_D$ instead of $\mathcal{L}$, while $\theta$ (i.e., parameters of $\text{GNN}_\theta$, $\text{MLP}_\theta$, and $p_\theta$) is meta-optimized based on $\mathcal{L}$.
	
	Moreover, it is noteworthy that our framework does not explicitly prevent collapse with extra operations (e.g., the negative samples used in contrastive learning~\cite{oord2018representation,he2020momentum}) when minimizing $\mathcal{L}_D$. Nevertheless, the loss design of our framework naturally avoids converging to a minimum regarding both $\theta$ and $\phi$ (e.g., a trivial constant representation). Different from BYOL~\cite{grill2020bootstrap} and BGRL~\cite{thakoor2021bootstrapped}, which utilize a momentum strategy, we propose two different losses (i.e., $\mathcal{L}_Y$ and $\mathcal{L}_D$) for meta-optimization regarding $\theta$ and $\phi$. As a result, the meta-optimization targets are different for $\theta$ and $\phi$ and thus will not cause collapse during meta-optimization. In addition, the collapse will also not occur during fine-tuning since only $\theta$ is updated while $\phi$ remains unchanged in this step.

	
	
	After meta-training on a specific number of meta-training tasks, we evaluate the performance of our framework X-FNC on the meta-test tasks, which are sampled from $\mathcal{C}_{test}$. 

	\begin{table*}[htbp]
		\setlength\tabcolsep{4.7pt}
	\small
		\centering
		\renewcommand{\arraystretch}{1.1}
		\caption{The overall few-shot node classification results (accuracy in \%) of X-FNC and baselines under different settings.}
        \vspace{-0.05in}
		\begin{tabular}{c||c|c|c|c|c|c||c|c|c|c|c|c}
			\hline
			Dataset&\multicolumn{6}{c||}{\texttt{DBLP}}&\multicolumn{6}{c}{\texttt{Amazon-E}}
			\\
			\hline
						Setting&\multicolumn{3}{c|}{5-way 3-shot}&\multicolumn{3}{c||}{10-way 3-shot}&\multicolumn{3}{c|}{5-way 3-shot}&\multicolumn{3}{c}{10-way 3-shot}\\
			\hline
			\# Labels per Class &5&10&20&5&10&20&5&10&20&5&10&20
			\\\hline
			\hline
PN&49.4\scriptsize{$\pm3.2$}&51.9\scriptsize{$\pm3.1$}&53.3\scriptsize{$\pm3.9$}&36.3\scriptsize{$\pm3.8$}&38.5\scriptsize{$\pm2.8$}&40.2\scriptsize{$\pm3.9$}&51.6\scriptsize{$\pm2.3$}&52.2\scriptsize{$\pm2.3$}&53.8\scriptsize{$\pm2.3$}&36.7\scriptsize{$\pm3.0$}&38.2\scriptsize{$\pm2.0$}&41.3\scriptsize{$\pm3.9$}\\\hline
MAML&50.9\scriptsize{$\pm3.1$}&51.8\scriptsize{$\pm1.8$}&56.1\scriptsize{$\pm2.1$}&39.4\scriptsize{$\pm2.3$}&44.3\scriptsize{$\pm2.0$}&45.4\scriptsize{$\pm3.1$}&48.8\scriptsize{$\pm2.4$}&49.4\scriptsize{$\pm3.3$}&53.9\scriptsize{$\pm2.7$}&39.0\scriptsize{$\pm3.2$}&40.3\scriptsize{$\pm3.2$}&41.5\scriptsize{$\pm3.2$}\\\hline
G-Meta&59.8\scriptsize{$\pm3.3$}&61.8\scriptsize{$\pm3.5$}&63.3\scriptsize{$\pm4.1$}&44.9\scriptsize{$\pm2.9$}&51.0\scriptsize{$\pm3.4$}&52.9\scriptsize{$\pm3.6$}&53.4\scriptsize{$\pm2.2$}&55.7\scriptsize{$\pm3.6$}&56.6\scriptsize{$\pm3.2$}&39.6\scriptsize{$\pm4.1$}&41.9\scriptsize{$\pm3.0$}&45.6\scriptsize{$\pm4.3$}\\\hline
GPN&58.6\scriptsize{$\pm3.8$}&62.5\scriptsize{$\pm2.8$}&66.9\scriptsize{$\pm4.3$}&50.6\scriptsize{$\pm3.9$}&52.7\scriptsize{$\pm2.4$}&54.6\scriptsize{$\pm3.4$}&56.0\scriptsize{$\pm4.1$}&60.7\scriptsize{$\pm4.7$}&63.0\scriptsize{$\pm2.3$}&42.1\scriptsize{$\pm4.8$}&45.8\scriptsize{$\pm3.3$}&52.1\scriptsize{$\pm4.8$}\\\hline
RALE&64.7\scriptsize{$\pm4.1$}&66.9\scriptsize{$\pm4.7$}&67.9\scriptsize{$\pm4.0$}&51.3\scriptsize{$\pm4.2$}&55.0\scriptsize{$\pm3.2$}&56.9\scriptsize{$\pm4.0$}&60.4\scriptsize{$\pm4.5$}&64.0\scriptsize{$\pm4.8$}&66.1\scriptsize{$\pm4.5$}&47.8\scriptsize{$\pm4.4$}&48.6\scriptsize{$\pm4.8$}&52.4\scriptsize{$\pm3.3$}\\\hline
X-FNC&\textbf{70.1\scriptsize{$\pm4.0$}}&\textbf{75.5\scriptsize{$\pm3.5$}}&\textbf{76.8\scriptsize{$\pm3.3$}}&\textbf{57.2\scriptsize{$\pm3.4$}}&\textbf{63.6\scriptsize{$\pm3.3$}}&\textbf{65.8\scriptsize{$\pm3.1$}}&\textbf{69.9\scriptsize{$\pm3.9$}}&\textbf{72.8\scriptsize{$\pm3.4$}}&\textbf{76.0\scriptsize{$\pm4.8$}}&\textbf{49.2\scriptsize{$\pm4.1$}}&\textbf{51.5\scriptsize{$\pm2.8$}}&\textbf{56.3\scriptsize{$\pm3.4$}}\\\hline

		\end{tabular}
		\label{tab:all_result}
		\vspace{-0.15cm}
	\end{table*}
	
	\begin{table*}[htbp]
		\setlength\tabcolsep{4.7pt}
	\small
		\centering
		\renewcommand{\arraystretch}{1.1}

		\begin{tabular}{c||c|c|c|c|c|c||c|c|c|c|c|c}
			\hline
			Dataset&\multicolumn{6}{c||}{\texttt{Cora-full}}&\multicolumn{6}{c}{\texttt{ogbn-arxiv}}\\\hline
			
						Setting&\multicolumn{3}{c|}{5-way 3-shot}&\multicolumn{3}{c||}{10-way 3-shot}&\multicolumn{3}{c|}{5-way 3-shot}&\multicolumn{3}{c}{10-way 3-shot}\\
			\hline
			\# Labels per Class &5&10&20&5&10&20&50&100&200&50&100&200
			\\\hline
			\hline
PN&45.5\scriptsize{$\pm2.7$}&48.1\scriptsize{$\pm3.6$}&48.9\scriptsize{$\pm3.8$}&28.2\scriptsize{$\pm3.8$}&31.6\scriptsize{$\pm3.3$}&34.4\scriptsize{$\pm2.5$}&39.1\scriptsize{$\pm2.5$}&40.8\scriptsize{$\pm3.7$}&42.6\scriptsize{$\pm3.1$}&23.1\scriptsize{$\pm3.6$}&24.4\scriptsize{$\pm3.1$}&27.7\scriptsize{$\pm3.5$}\\\hline
MAML&46.9\scriptsize{$\pm2.6$}&48.6\scriptsize{$\pm3.0$}&49.2\scriptsize{$\pm2.7$}&32.7\scriptsize{$\pm2.5$}&33.2\scriptsize{$\pm2.3$}&35.8\scriptsize{$\pm1.9$}&41.0\scriptsize{$\pm2.4$}&41.9\scriptsize{$\pm1.9$}&43.1\scriptsize{$\pm3.4$}&23.2\scriptsize{$\pm2.1$}&25.4\scriptsize{$\pm3.1$}&28.0\scriptsize{$\pm3.2$}\\\hline
G-Meta&57.7\scriptsize{$\pm3.9$}&58.7\scriptsize{$\pm3.6$}&59.8\scriptsize{$\pm2.6$}&41.7\scriptsize{$\pm3.3$}&42.0\scriptsize{$\pm3.0$}&43.8\scriptsize{$\pm2.7$}&43.5\scriptsize{$\pm3.6$}&44.7\scriptsize{$\pm2.9$}&46.5\scriptsize{$\pm4.4$}&27.4\scriptsize{$\pm4.4$}&29.0\scriptsize{$\pm2.8$}&29.9\scriptsize{$\pm2.5$}\\\hline
GPN&54.6\scriptsize{$\pm2.8$}&55.2\scriptsize{$\pm3.6$}&57.7\scriptsize{$\pm4.2$}&38.4\scriptsize{$\pm2.8$}&40.2\scriptsize{$\pm2.9$}&42.0\scriptsize{$\pm4.5$}&46.6\scriptsize{$\pm3.4$}&47.1\scriptsize{$\pm3.9$}&48.4\scriptsize{$\pm2.9$}&26.1\scriptsize{$\pm2.6$}&30.9\scriptsize{$\pm3.6$}&33.5\scriptsize{$\pm3.5$}\\\hline
RALE&58.2\scriptsize{$\pm2.8$}&59.3\scriptsize{$\pm4.1$}&63.1\scriptsize{$\pm3.9$}&38.1\scriptsize{$\pm4.2$}&43.4\scriptsize{$\pm2.8$}&44.0\scriptsize{$\pm4.5$}&49.3\scriptsize{$\pm3.0$}&51.4\scriptsize{$\pm3.9$}&52.5\scriptsize{$\pm4.6$}&30.4\scriptsize{$\pm2.5$}&31.7\scriptsize{$\pm3.3$}&33.9\scriptsize{$\pm4.8$}\\\hline
X-FNC&\textbf{62.9\scriptsize{$\pm4.5$}}&\textbf{68.0\scriptsize{$\pm3.7$}}&\textbf{69.2\scriptsize{$\pm4.6$}}&\textbf{43.7\scriptsize{$\pm4.8$}}&\textbf{45.6\scriptsize{$\pm4.4$}}&\textbf{47.7\scriptsize{$\pm4.5$}}&\textbf{54.6\scriptsize{$\pm2.6$}}&\textbf{56.7\scriptsize{$\pm4.0$}}&\textbf{58.7\scriptsize{$\pm4.1$}}&\textbf{33.3\scriptsize{$\pm3.9$}}&\textbf{35.7\scriptsize{$\pm4.4$}}&\textbf{39.8\scriptsize{$\pm2.4$}}\\\hline

		\end{tabular}
\vspace{-0.1in}
	\end{table*}

    \section{Experimental Evaluations}

\subsection{Datasets}

	\begin{table}[htbp]
	\setlength\tabcolsep{4pt}
		\small
		\centering
		\renewcommand{\arraystretch}{1.1}

		\caption{Statistics of four node classification datasets. }
        \vspace{-0.1in}
		\begin{tabular}{c|c|c|c|c}
		\hline
        \textbf{Dataset}&\# Nodes & \# Edges & \# Features & Class Split\\
        \hline
        \texttt{Amazon-E}&42,318& 43,556&8,669& 90/37/40 \\
        \texttt{DBLP}&40,672&288,270&7,202&80/27/30 \\
        \texttt{Cora-full}&19,793&65,311&8,710&25/20/25 \\
        \texttt{ogbn-arxiv}&169,343&1,166,243&128&15/5/20 \\
        
        \hline
		\end{tabular}
        \vspace{-0.1in}
		\label{tab:statistics}
	\end{table}

To evaluate the performance of X-FNC on few-shot node classification with extremely weak supervision, we conduct experiments on four prevalent real-world graph datasets: \texttt{Amazon-E}~\cite{mcauley2015inferring}, \texttt{DBLP}~\cite{tang2008arnetminer}, \texttt{Cora-full}~\cite{bojchevski2018deep}, and \texttt{ogbn-arxiv}~\cite{hu2020open}. Each dataset is a graph and consists of a considerable number of node classes to ensure that the meta-test tasks contain a variety of classes for a more comprehensive evaluation. 
Specifically, we obtain \texttt{Amazon-E} and \texttt{DBLP} datasets from~\cite{ding2020graph}. \texttt{Cora-full} and \texttt{ogbn-arxiv} are from the corresponding source. Then we conduct experiments on these datasets under the extremely weak supervision setting. In particular, we choose three different settings: 5/10/20 labels per class. In other words, each meta-training class only consists of 5/10/20 labeled nodes (50/100/200 for \texttt{ogbn-arxiv} due to the large size of the graph), where the total labeled nodes are approximately 1\%/2\%/4\% of nodes on the graph. It is noteworthy that we randomly select these labeled nodes from the training classes in the original datasets.
The detailed statistics of these datasets are summarized in Table~\ref{tab:statistics}, where the class split setting denotes the number of classes used for training/validation/test.

\subsection{Experimental Settings}
To achieve a comparison of X-FNC with competitive baselines, we conduct experiments with the state-of-the-art few-shot node classification methods. \textbf{Prototypical Networks (PN)}~\cite{snell2017prototypical} and \textbf{MAML}~\cite{finn2017model} are conventional few-shot methods, and we apply them on graph data. 
\textbf{G-Meta}~\cite{huang2020graph}, \textbf{GPN}~\cite{ding2020graph}, and \textbf{RALE}~\cite{liu2021relative} are recently proposed studies on few-shot node classification. 

During meta-training, we randomly sample $\mathcal{T}_{train}$ meta-training tasks from meta-training classes for model optimization. Here the support set and the query set in each meta-task will only be sampled from labeled nodes (i.e., 5/10/20 labeled nodes in each class). 
Then during meta-test, we evaluate the model on a series of randomly sampled meta-test tasks from the entire node set of meta-test classes. 
The final averaged classification accuracy on meta-test tasks will be used as the evaluation metric. 
For the Poisson Label Propagation module, we set the number of label propagation steps $T_l$ as 10. The number of randomly sampled nodes $R$ to construct the subgraph for label propagation is set as 10. The scaling parameter of $\mathbf{A}''$ is set as 100, and the hyper-parameter $\lambda$ is set as 0.5. The number of selected pseudo-labeled nodes $M$ is 20. For IB fine-tuning, the number of fine-tuning steps $T$ is 40. The mask rate $\gamma$ is set as 0.1. The learning rate $\alpha$ during fine-tuning is 0.1. The meta-learning rates $\beta_1$ and $\beta_2$ during meta-optimization are set as 0.005.
The trade-off hyper-parameter $\beta$ is set as 1. The hidden sizes of $\text{GNN}_\theta$ and $\text{GNN}_\phi$ are both 64. The hidden size of $\text{MLP}_\theta$ is 64 while the hidden sizes of the two MLP layers in $p_\theta$ are 128 and 64, respectively. The number of training epochs $T_{train}$ is 5,000, and the number of meta-test tasks $T_{test}$ is 500. The dropout rate is 0.5. The query set size $|\mathcal{Q}|$ is 10. Our code can be found at \href{https://github.com/SongW-SW/X-FNC}{https://github.com/SongW-SW/X-FNC}.

\subsection{Performance Comparison}
Table~\ref{tab:all_result} presents the performance comparison of our framework X-FNC and all baselines on few-shot node classification with extremely weak supervision. Specifically, we choose two different few-shot settings to obtain a more comprehensive comparison: 5-way 3-shot and 10-way 3-shot. We use the average classification accuracy over 10 repetitions as the evaluation metric. From Table~\ref{tab:all_result}, we can have the following observations: (1) Our framework X-FNC achieves the best results compared with all other baselines in all datasets. The performance also consistently outperforms other baselines under different settings, which validates the superiority of X-FNC on few-shot node classification with extremely weak supervision. (2) When the number of labels per class decreases from 20 to 5, X-FNC has the least performance drop compared with other baselines. The main reason is that X-FNC obtains pseudo-labeled nodes via Poisson Label Propagation to alleviate the under-generalizing problem with extremely weak supervision. (3) The performance improvement of X-FNC over other baselines is slightly larger on \texttt{DBLP}. This is due to the fact that \texttt{DBLP} has a larger average node degree, which helps improve the pseudo-labeling accuracy during meta-training for better performance. (4) When the value of $N$ increases (i.e., more classes in each meta-task), all methods encounter a significant performance drop, since query nodes are classified from a larger class set in each meta-task. Nevertheless, under the extremely limited setting, X-FNC consistently outperforms other baselines. It is because X-FNC can better extract decisive information for classification with a larger value of $N$ via IB fine-tuning.

\subsection{Ablation Study}
\begin{figure}[!t]
		\centering
		\subfigure{
\includegraphics[width=0.23\textwidth]{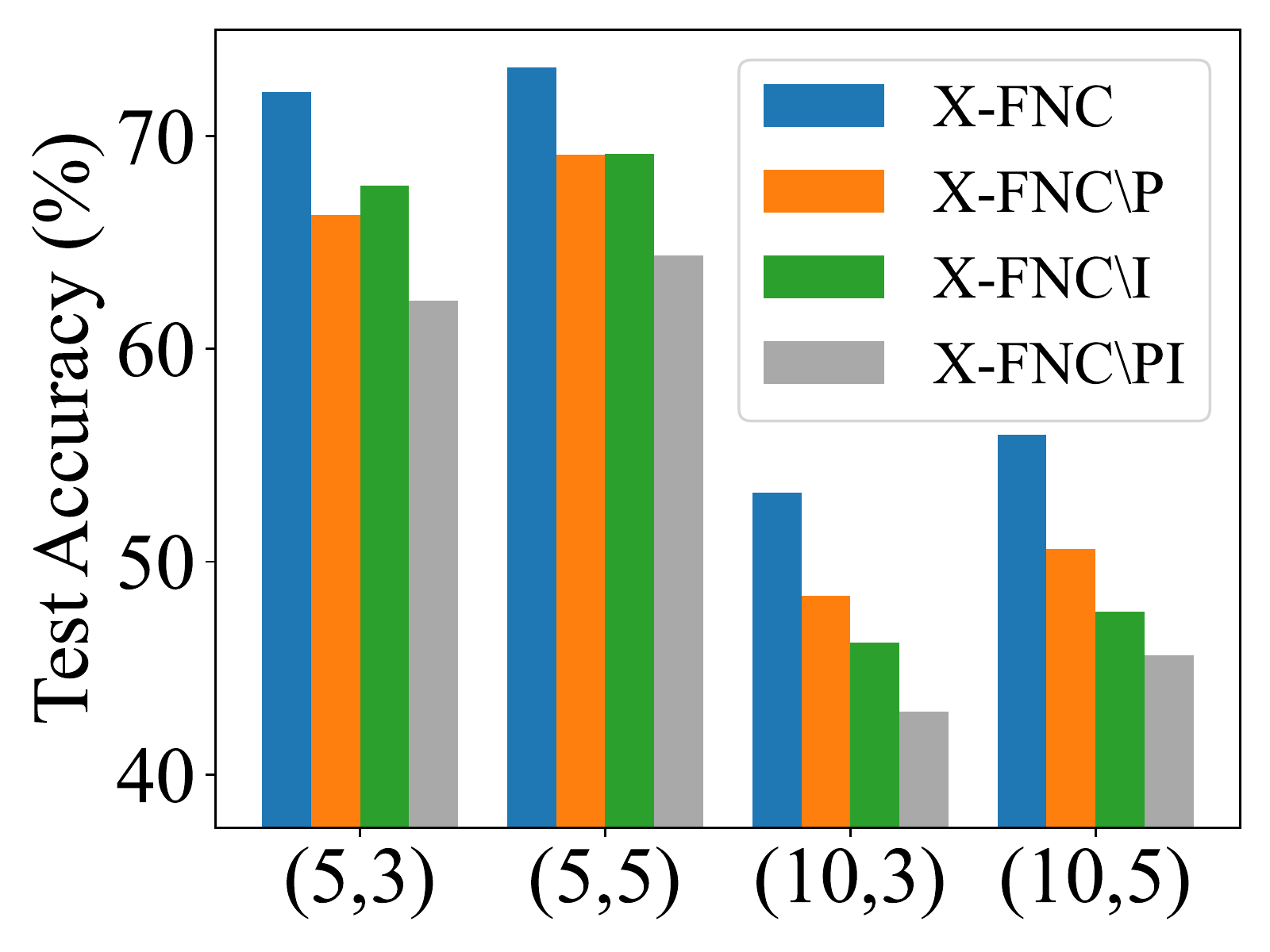}}
	\subfigure{
\includegraphics[width=0.23\textwidth]{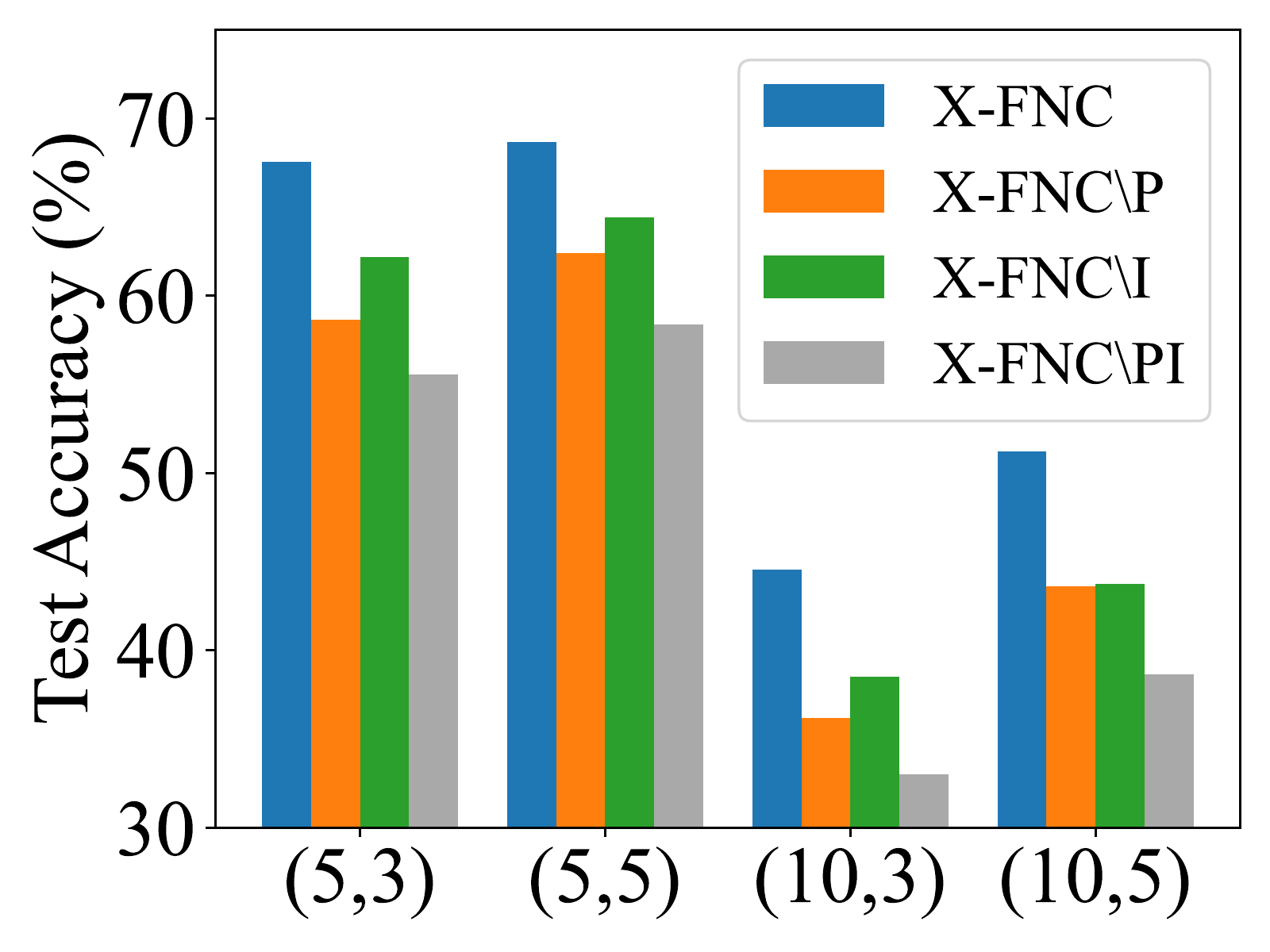}}
\vspace{-0.2in}
		\caption{Ablation study on our framework on \texttt{Amazon-E} (left) and \texttt{Cora-full} (right) in the $N$-way $K$-shot setting ($N$, $K$).}
		\label{fig:ablation}
\vspace{-0.15in}
	\end{figure}

We conduct an ablation study on \texttt{Amazon-E} and \texttt{Cora-full} to evaluate the effectiveness of different components in our framework X-FNC (similar results on other datasets). Specifically, we compare X-FNC with three degenerate versions: (a) X-FNC without pseudo-labeling (X-FNC\textbackslash P); (b) X-FNC without IB-based fine-tuning (X-FNC\textbackslash I); (c) without both (X-FNC\textbackslash PI). More specifically, X-FNC\textbackslash P removes the pseudo-labeling process such that the support set only consists of the given labeled nodes. X-FNC\textbackslash I replaces the IB fine-tuning process with a simple classifier during fine-tuning, while X-FNC\textbackslash PI combines the two variants. From Fig.~\ref{fig:ablation}, we can obtain several observations. First, our framework outperforms all other variants, which further validates that each module plays an important role in few-shot node classification with extremely weak supervision. Second, removing the Poisson Learning module deteriorates the performance on \texttt{Cora-full} more than that on \texttt{Amazon-E}. The reason is that \texttt{Cora-full} consists of significantly fewer meta-training classes than \texttt{Amazon-E}, and obtaining pseudo-labeled nodes becomes more crucial in this scenario. Third, without IB fine-tuning, the performance drops more significantly on 10-way settings than 5-way settings. The result further indicates that IB fine-tuning is critical for model generalization to meta-test classes, especially when each meta-task includes more classes.

		\begin{figure}[!t]
		\centering
		\subfigure{
\includegraphics[width=0.23\textwidth]{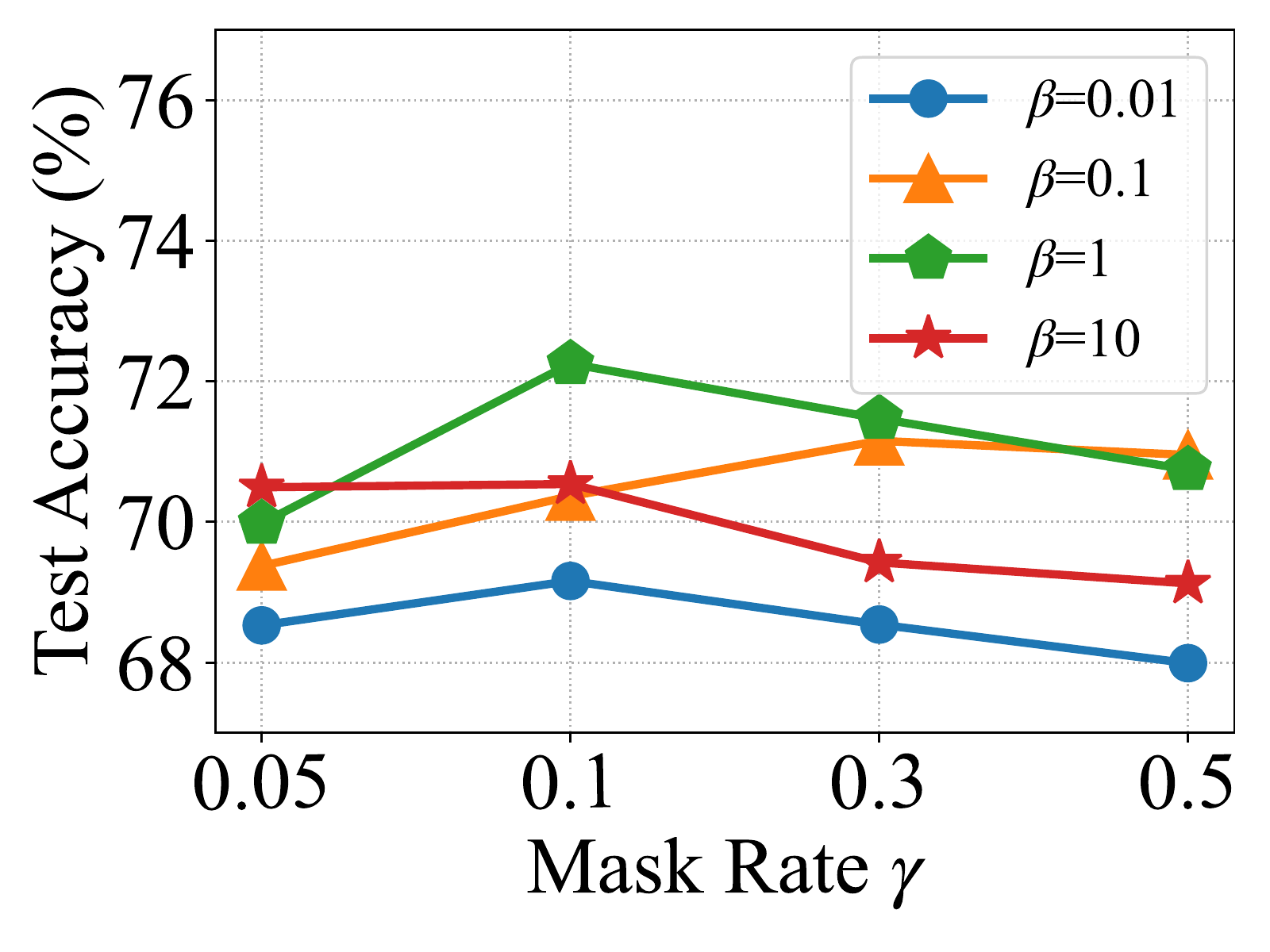}}
	\subfigure{
\includegraphics[width=0.23\textwidth]{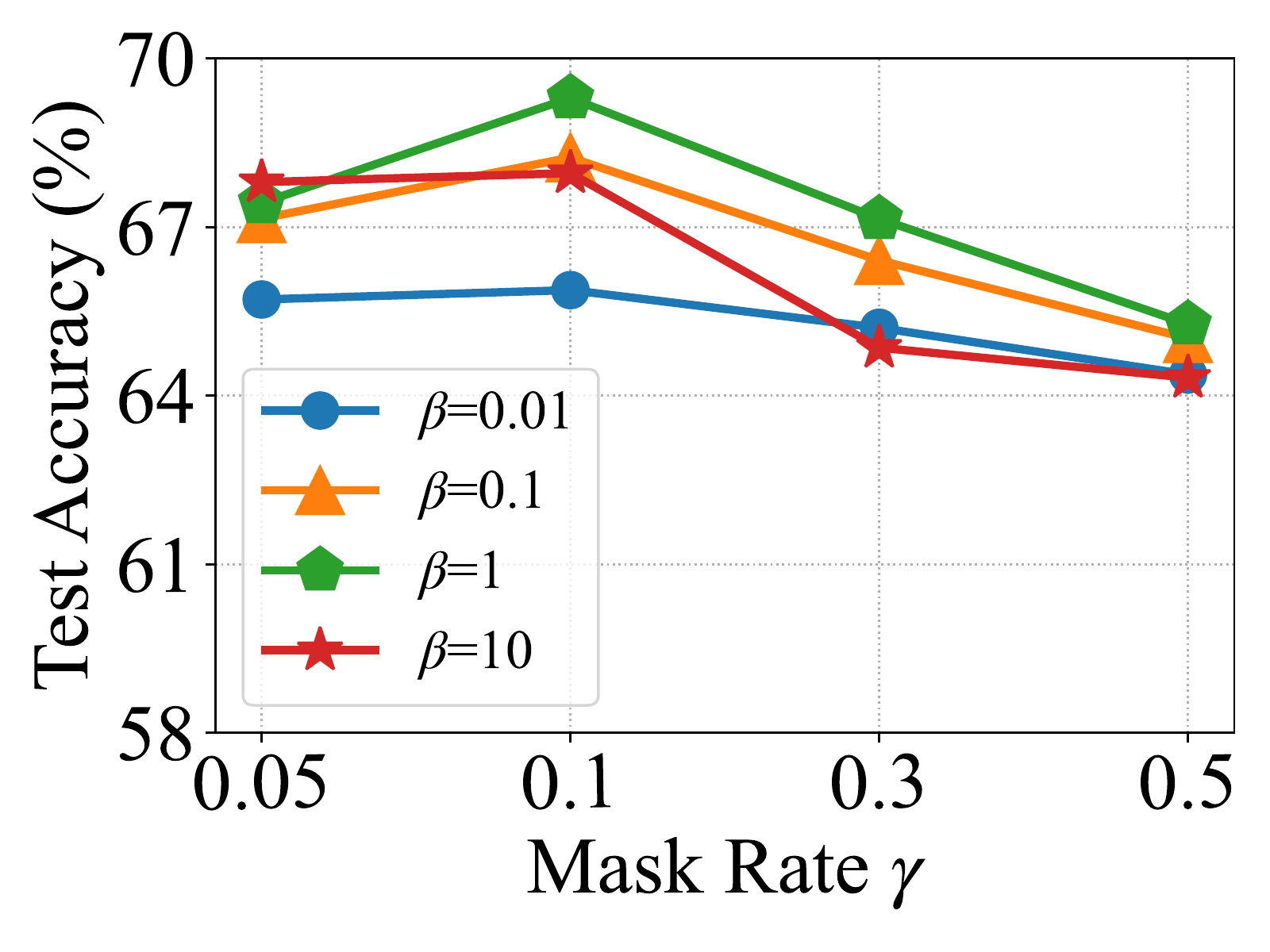}}
\vspace{-0.2in}
		\caption{Results of our framework on \texttt{Amazon-E} (left) and \texttt{Cora-full} (right) with different mask rates.}
		\label{fig:Q}
	\vspace{-0.15in}
	\end{figure}

\subsection{Effect of Loss $\mathcal{L}_D$}
In this part, we conduct experiments to study the effect of the loss $\mathcal{L}_D$ in IB fine-tuning, which is calculated according to Eq. (\ref{eq:ld}) with two hyper-parameters (i.e., the loss weight $\beta$ and the mask rate $\gamma$). 
Specifically, in X-FNC, $\beta$ represents the level of attention the model pays to the irrelevant local structures for classification on a specific node. According to the IB principle, with a higher loss weight $\beta$, the model will focus more on filtering out irrelevant information for classification while less on extracting decisive classification information. On the other hand, the mask rate $\gamma$ represents the approximate ratio of irrelevant information in the local structure of each node, which should be adjusted according to different datasets. To demonstrate the joint impact of these two hyper-parameters, we present the results with different values of $\beta$ and $\gamma$ on \texttt{Amazon-E} and \texttt{Cora-full}. From Fig.~\ref{fig:Q}, we can observe that the mask rate of 0.1 generally provides better performance than other values. This is mainly because a small mask rate can be insufficient to filter out irrelevant structural information, while a larger mask rate can result in the loss of helpful information in local structures. Moreover, the performance drop on \texttt{Cora-full} is slightly larger than \texttt{Amazon-E}. The reason is that in \texttt{Cora-full}, the average node degree is significantly larger than \texttt{Amazon-E}. As a result, the graph structure encodes more decisive information for classification, which is more easily impacted by a large mask rate.

\subsection{Random Sampling in Pseudo-labeling}
\begin{figure}[!t]
		\centering
		\subfigure{
\includegraphics[width=0.23\textwidth]{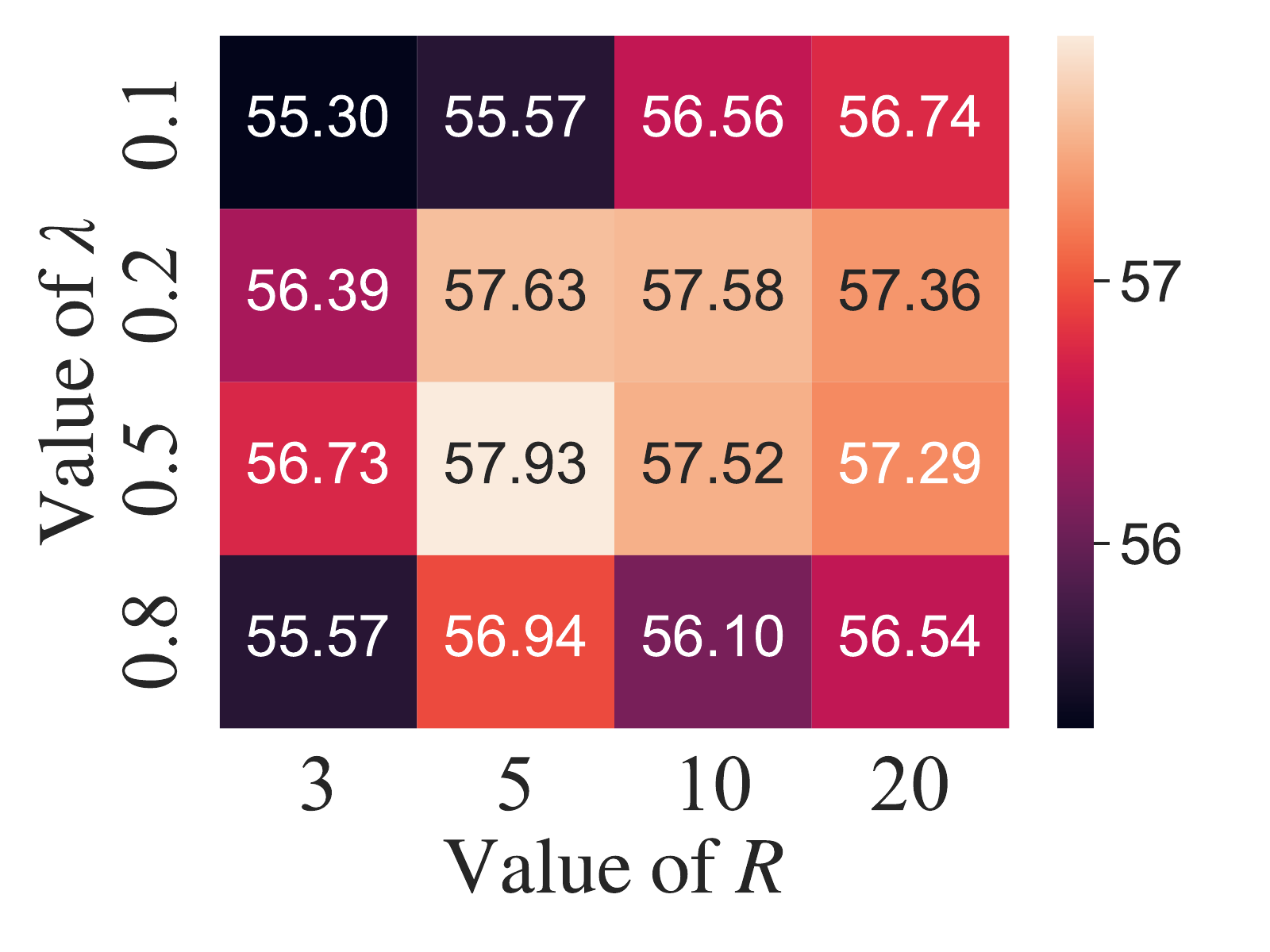}}
	\subfigure{
\includegraphics[width=0.23\textwidth]{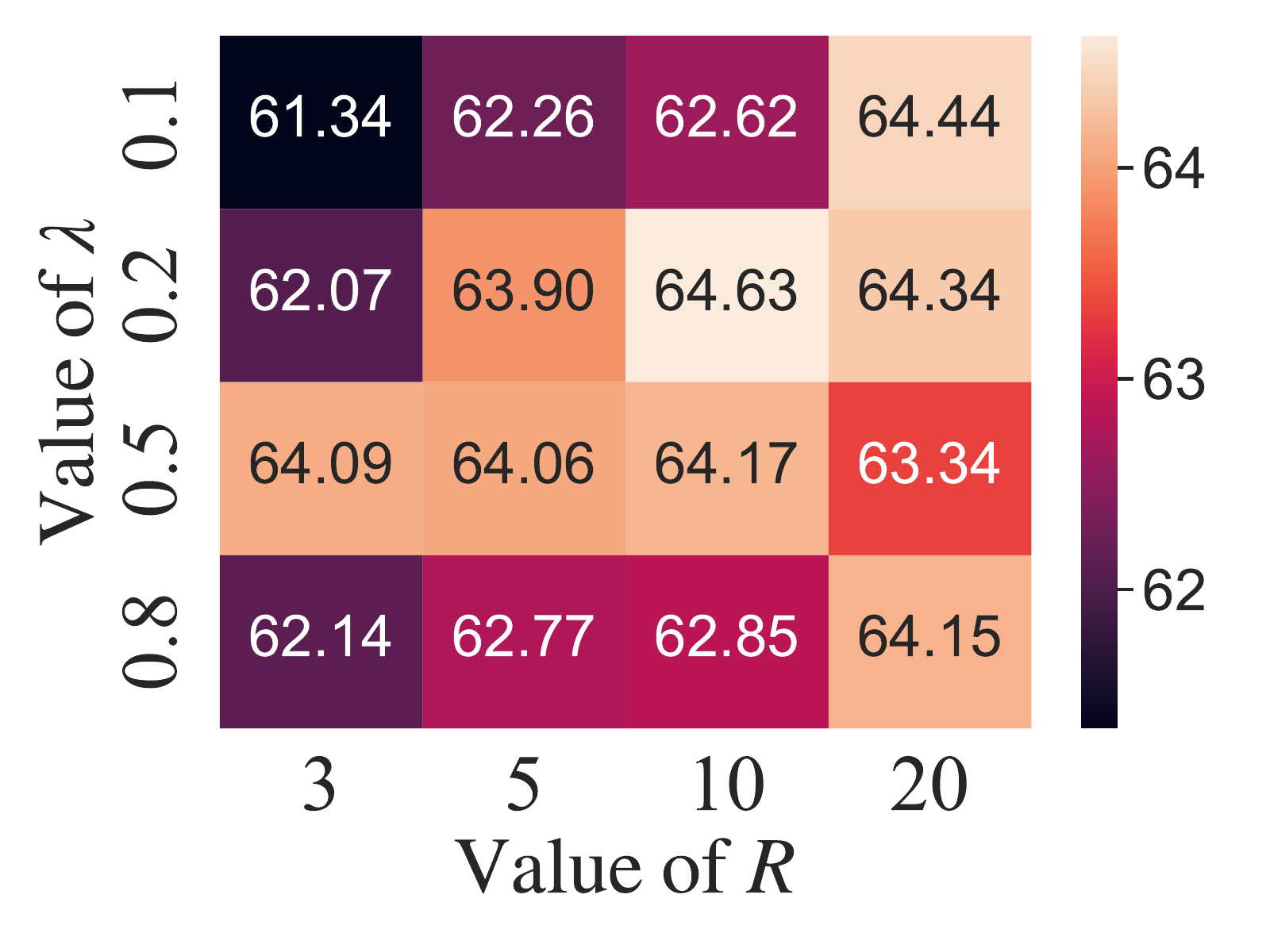}}
\vspace{-0.2in}
		\caption{Results of pseudo-labeling accuracy (in \%) on \texttt{Amazon-E} (left) and \texttt{Cora-full} (right) with different $\lambda$ and $R$.}
		\label{fig:NK}
	\vspace{-0.15in}
	\end{figure}

In this part, we study the impact of factors that affect the pseudo-labeling accuracy during label propagation.
Specifically, the sample number $R$ controls the ratio of random unlabeled nodes in the constructed subgraph during pseudo-labeling. 
On the other hand, the distance-based adjacency matrix $\mathbf{A}''$ acts as flexible connections between randomly sampled nodes and the limited labeled nodes (i.e., support nodes). Hence, we also adjust the values of $R$ and the scaling hyper-parameter $\lambda$ to evaluate their influence. From Fig.~\ref{fig:NK}, we observe that the two parameters affect the pseudo-labeling accuracy differently. In particular, increasing the number of randomly sampled nodes $R$ will first increase pseudo-labeling accuracy and then keep it stable. It is because a more complex structure of the constructed subgraph can help the label propagation process. In addition, a higher $\lambda$ (i.e., the scaling hyper-parameter for $\mathbf{A}''$) first increases the pseudo-labeling accuracy while later deteriorating the accuracy. The reason is that with a higher value of $\lambda$, the model will focus more on label propagation to random nodes instead of neighbors of support nodes. As a result, a higher $\lambda$ can help discover more nodes that share the same classes with support nodes.


    \section{Conclusion}
    In this paper, we study the problem of few-shot node classification with extremely weak supervision, which focuses on predicting labels for nodes in meta-test classes while utilizing extremely limited labeled nodes for meta-training. Furthermore, to tackle the challenges caused by extremely limited labeled nodes, we propose an innovative framework X-FNC to obtain pseudo-labeled nodes via Poisson Learning and conduct fine-tuning based on the IB principle. As a result, our framework can expand the support set in each meta-task to alleviate the problem of under-generalizing while filtering out irrelevant information for classification to avoid over-fitting. We conduct extensive experiments on four node classification datasets with extremely weak supervision, and the results validate the superiority of our framework over other state-of-the-art baselines. 

\section*{Acknowledgements} This work is supported by the National Science Foundation under grants IIS-2006844, IIS-2144209,
IIS-2223769, CNS-2154962, and BCS-2228534, the JP Morgan Chase Faculty Research Award, the Cisco Faculty Research Award, the Jefferson Lab subcontracts JSA-22-D0311 and JSA-23-D0163, the Commonwealth Cyber Initiative awards HV-2Q23-003 and VV-1Q23-007, the 4-VA Collaborative Research grant, and the UVA 3Cavaliers Seed Research Grant.


\end{document}